\documentclass[runningheads]{llncs}
\usepackage[T1]{fontenc}
\usepackage{graphicx}
\usepackage{booktabs}
\usepackage{amssymb}
\usepackage[misc]{ifsym}

\usepackage{mwe}

\usepackage{multirow}
\usepackage[dvipsnames]{xcolor}
\usepackage{bm}
\usepackage{tablefootnote}
\usepackage[inline]{enumitem}
\usepackage[normalem]{ulem}
\usepackage{subfigure}
\usepackage{comment}
\usepackage{pifont}

\usepackage{xcolor}
\usepackage{pifont}
\usepackage{multirow}
\usepackage{makecell}

\usepackage{fontawesome5}

\newcommand{\cmark}{\ding{51}}
\newcommand{\xmark}{\ding{55}}%

\begin{document}

\title{Datum-wise Transformer for Synthetic Tabular Data Detection in the Wild}

\titlerunning{Datum-wise Transformer Transformer for Synthetic Tabular Data Detection}


\author{G. Charbel N. Kindji\inst{1,2} \and
Elisa Fromont\inst{2}\orcidID{0000-0003-0133-349} \and
Lina M. Rojas-Barahona\inst{1}\orcidID{0009-0009-8439-4695} \and
Tanguy Urvoy \inst{1}}

\institute{Orange Labs Lannion \\
\email{first.last@orange.com} \and
Université de Rennes, CNRS, Inria, IRISA UMR 6074.\\
\email{first.last@irisa.fr}}

\authorrunning{Kindji et al.}

\maketitle              

\begin{abstract}
The growing power of generative models raises major concerns about the authenticity of published content. To address this problem, several synthetic content detection methods have been proposed for uniformly structured media such as image or text. However, little work has been done on the detection of synthetic tabular data, despite its importance in industry and government. This form of data is complex to handle due to the diversity of its structures: the number and types of the columns may vary wildly from one table to another. We tackle the tough problem of detecting synthetic tabular data "in the wild", \textit{i.e.} when the model is deployed on table structures it has never seen before. We introduce a novel datum-wise transformer architecture and show that it outperforms existing models. Furthermore, we investigate the application of domain adaptation techniques to enhance the effectiveness of our model, thereby providing a more robust data-forgery detection solution.
\end{abstract}

\section{Introduction}

Over the past five years, deep learning-based generative models have surged in popularity \cite{generative24}, raising significant concerns about their potential misuse \cite{marchal2024generativeaimisusetaxonomy}, including opinion manipulation, fraud, and harassment. In response, numerous detection methods have been developed for uniformly structured media such as images and text \cite{fakeportrait20,faketext25}. However, detecting synthetic tabular data remains largely underexplored, despite its critical role in industries such as healthcare, finance, and government.
We focus on this problem and frame the detection of synthetic tabular data as a binary classification problem. Unlike other data modalities, tabular data poses unique challenges due to its heterogeneity: varying numbers of features, diverse feature types and distributions, as well as differences in table sizes. Additionally, an effective synthetic tabular data detector must be \textit{table-agnostic}, meaning it should function independently of a fixed table structure. This requirement disqualifies most state-of-the-art tabular predictors, including \cite{breiman2001random,chen2016xgboost,prokhorenkova2018catboost}, as well as recent transformer-based models tailored to specific tabular structures \cite{huang2020tabtransformer,arik2021tabnet,somepalli2022saint}.

As introduced in \cite{KindjiIDA2025}, the detection of synthetic tabular data can be categorized into three levels, each representing a different degree of "wildness":
\begin{enumerate}
    \item \textit{Same-table} detection: Identifying synthetic data within a single table structure, as seen in \textit{Classifier Two-Sample Test} (C2ST)~\cite{lopez2016revisiting,kindji2024hoodtabulardatageneration}. In this setting, the detector does not need to be table-agnostic.
    \item \textit{Cross-table} detection: Identifying synthetic data across multiple tables (e.g., detecting real and synthetic rows in both \textit{Adult} and \textit{Insurance} tables). This setup requires a table-agnostic detector, similar to text-based detectors that generalize across different tables within a predefined corpus.
    \item \textit{Cross-table shift} detection: Handling deployment scenarios where the tables encountered at inference differ from those seen during training (e.g., training on \textit{Adult} and testing on \textit{Insurance}).
\end{enumerate}
Each level presents increasing challenges. We focus here on the \textit{cross-table shift} scenario which is both the most realistic and the most demanding one.
We propose a novel table-agnostic transformer architecture for synthetic tabular data detection under \textit{cross-table shift}. To ensure comparability, we follow the methodology recently proposed in~\cite{KindjiIDA2025}, in which several baseline detectors were introduced.  We show that our model significantly outperforms these baselines, further improving with the integration of a domain adaptation strategy~\cite{ganin2015unsupervised}. Moreover, our architecture, being both table-agnostic and invariant to column permutation, holds potential for future applications beyond synthetic data detection.

The next sections are structured as follows. We present the related work in Section~\ref{s:related_work} followed by the detailed description of our model in Section~\ref{s:table_agnostic_eq_transf}. We then present our experimental setup and results in Section~\ref{s:experiments}. Finally, we conclude and outline future research directions in Section~\ref{s:conclusion}.

\section{Related Work}
\label{s:related_work}

\begin{table}
\centering
\caption{Table encoding properties. \textbf{Input}: whether the model takes \textit{full tables} (\textit{table}) as {input} or tackle them \textit{row-by-row} (\textit{row}). \textbf{Cross-table deployment}: whether the model can be deployed on different types of table or on a \textit{single-table} (\textit{single}) type. 
\textbf{Internal data-format}: whether the feature encoding, is \textit{type-specific} (e.g. numerical) or \textit{text-based} (\textit{text}). \textbf{Independent feature encoding}: whether the feature encoding stage is global or independent for each attribute of the table.
\textbf{Permutation Invariant}: whether the predictions are invariant by permutation of the columns of the table.}
\label{tab:relwork}
\renewcommand\theadfont{\bfseries}
\begin{tabular}{|c|c|c|c|c|c|}
\hline
\thead{Method}    & \thead{Input}            & \multicolumn{1}{c|}{\thead{Cross-table \\deployment}} & \thead{Internal \\data format} & 
\multicolumn{1}{c|}{\thead{Indep.\\ feat. embd.}} & \thead{Permut. \\inv.} \\ \hline
         
TaBERT~\cite{yin20acl}    & table                    & single                                    & text                 & \xmark                                                 & \xmark                                              \\ \hline

\multicolumn{1}{|c|}{TABBIE~\cite{iida-etal-2021-tabbie}} & \multicolumn{1}{c|}{table}  & \multicolumn{1}{c|}{single}                           & \multicolumn{1}{c|}{text} & \multicolumn{1}{c|}{\xmark}                              & \multicolumn{1}{c|}{\cmark}      \\ \hline

\multicolumn{1}{|c|}{TAPAS~\cite{herzig-etal-2020-tapas}}     & \multicolumn{1}{c|}{table}             & \multicolumn{1}{c|}{single}               & \multicolumn{1}{c|}{text}                 & \multicolumn{1}{c|}{\xmark}                            & \multicolumn{1}{c|}{\xmark}                    \\ \hline
\multicolumn{1}{|c|}{TAPEX~\cite{liu2022tapex}}   & \multicolumn{1}{c|}{table}             & \multicolumn{1}{c|}{single}               & \multicolumn{1}{c|}{text}                 & \multicolumn{1}{c|}{\xmark}                            & \multicolumn{1}{c|}{\xmark}                    \\ \hline

\multicolumn{1}{|c|}{STUNT~\cite{nam2023stunt}}    & \multicolumn{1}{c|}{table}             & \multicolumn{1}{c|}{single}               & \multicolumn{1}{c|}{type-specific}               & \multicolumn{1}{c|}{\cmark}                           & \multicolumn{1}{c|}{\cmark}                    \\ \hline
\multicolumn{1}{|c|}{PTaRL~\cite{PtarlYe2024}}     & \multicolumn{1}{c|}{table}             & \multicolumn{1}{c|}{single}               & \multicolumn{1}{c|}{type-specific}               & \multicolumn{1}{c|}{\cmark}                           & \multicolumn{1}{c|}{\cmark}                   \\ \hline

\multicolumn{1}{|c|}{Xtab~\cite{XtabZhu2023}}      & \multicolumn{1}{c|}{row}             & \multicolumn{1}{c|}{single}               & \multicolumn{1}{c|}{type-specific}               & \multicolumn{1}{c|}{\cmark}                           & \multicolumn{1}{c|}{\cmark}                   \\ \hline
\multicolumn{1}{|c|}{Unitabe~\cite{yang2024unitabe}}   & \multicolumn{1}{c|}{row}             & \multicolumn{1}{c|}{single}               & \multicolumn{1}{c|}{type-specific}               & \multicolumn{1}{c|}{\cmark}                           & \multicolumn{1}{c|}{\cmark}                   \\ \hline
\multicolumn{1}{|c|}{Transtab~\cite{wang2022transtab}}  & \multicolumn{1}{c|}{row}             & \multicolumn{1}{c|}{multi-tables~\tablefootnote{Tables with overlaping columns}}                & \multicolumn{1}{c|}{type-specific}               & \multicolumn{1}{c|}{\cmark}                           & \multicolumn{1}{c|}{\cmark}                   \\ \hline
\multicolumn{1}{|c|}{PORTAL~\cite{spinaci2024portal}}    & \multicolumn{1}{c|}{row}             & \multicolumn{1}{c|}{single}               & \multicolumn{1}{c|}{type-specific}               & \multicolumn{1}{c|}{\cmark}                           & \multicolumn{1}{c|}{\cmark}                   \\ \hline

\multicolumn{1}{|c|}{CARTE~\cite{kimcarte}}    & \multicolumn{1}{c|}{row}             & \multicolumn{1}{c|}{single}               & \multicolumn{1}{c|}{type-specific}               & \multicolumn{1}{c|}{\cmark}                           & \multicolumn{1}{c|}{\cmark}                   \\ \hline

\multicolumn{1}{|c|}{SwitchTab~\cite{switchTabJing2024}}    & \multicolumn{1}{c|}{table}             & \multicolumn{1}{c|}{single}               & \multicolumn{1}{c|}{type-specific}               & \multicolumn{1}{c|}{\xmark}                           & \multicolumn{1}{c|}{\cmark}                   \\ \hline

\multicolumn{1}{|c|}{TabPFN~\cite{Hollmann2025TabPFN}}    & \multicolumn{1}{c|}{table}             & \multicolumn{1}{c|}{single}               & \multicolumn{1}{c|}{type-specific}               & \multicolumn{1}{c|}{\cmark}                           & \multicolumn{1}{c|}{\cmark}                   \\ \hline

\multicolumn{1}{|c|}{TabuLa-8B~\cite{gardner2024large}} & \multicolumn{1}{c|}{row}             & \multicolumn{1}{c|}{single}               & \multicolumn{1}{c|}{text}                 & \multicolumn{1}{c|}{\xmark}                            & \multicolumn{1}{c|}{\cmark}                   \\ \hline
\multicolumn{1}{|c|}{STab~\cite{hajiramezanali2022stab}}      & \multicolumn{1}{c|}{row}             & \multicolumn{1}{c|}{single}               & \multicolumn{1}{c|}{type-specific}               & \multicolumn{1}{c|}{\xmark}                            & \multicolumn{1}{c|}{\cmark}                   \\ \hline


Flat Text~\cite{KindjiIDA2025} & row                   & mixed tables                                    & text                 & \xmark                                                 & \xmark                                            \\ \hline

Datum-wise (ours)    & row                   & mixed tables                                    & text                 & \cmark                                                & \cmark                                            \\ \hline
\end{tabular}
\end{table}

Research involving tabular data is increasingly shifting towards the development of foundation models \cite{yin20acl,iida-etal-2021-tabbie,herzig-etal-2020-tapas,liu2022tapex}, akin to those established for text and images. These models function as generic embedding machines, capable of generalizing across diverse tasks involving a wide range of input tables. Indeed, tabular foundation models are required to be table-agnostic, at least for the cross-table pretraining phase.
Their performance is however primarily evaluated on table-specific regression and classification tasks. In the following, we present some of the foundation models that can serve as competitors for our proposed synthetic data detection solution. The key properties of these models are summarized in Table~\ref{tab:relwork}. In particular, we differentiate the solutions by, the type of input they take into account (whole tables or rows), if they have a built-in mechanisms to process tabular data of varying structures, how they deal with feature encoding ("everything is text" or type-specific and global or "by-attribute") and whether the predictive result is the same if the table columns are permuted. Note that existing models such as TabPFN~\cite{Hollmann2025TabPFN} or TabuLa-8B~\cite{gardner2024large} explicitly introduce robustness to column permutation during their pretraining phase, while others may be more or less sensitive to this order depending on their (pre)training methods. We focus on the invariance in downstream tasks.

In order to be table-agnostic, most tabular data encoders rely on a textual representation of the table.
Several models, such as TaBERT~\cite{yin20acl} and TABBIE~\cite{iida-etal-2021-tabbie}, are built upon the well-known BERT~\cite{Devlin2019BERTPO} transformer encoder for text data. TaBERT jointly learns tabular data representations alongside contextual information from natural language (NL) sentences. During its pretraining phase, TaBERT selects $K$ rows from the table based on their relevance to the NL context provided. The authors propose two pretrained versions: one with $K=1$ and another with $K=3$, corresponding to the selection of one or three rows, respectively. Regarding TABBIE, it processes tables directly without requiring additional context.
TAPAS~\cite{herzig-etal-2020-tapas} and TAPEX~\cite{liu2022tapex} are designed for question answering on small (Wikipedia-like) tables. Similar to TaBERT, TAPAS is based on BERT and processes both NL sentences and tabular data. TAPEX, on the other hand, pretrains a Large Language Model (LLM) to execute SQL queries and finetunes it for downstream tasks using tables combined with NL inputs.
 
Some models rely on \textit{type-specific} feature encodings instead of text, especially for numerical features. These encoding must be generic enough for the model to remain table-agnostic, at least during its pretraining phase.
These include Xtab~\cite{XtabZhu2023}, UniTabE~\cite{yang2024unitabe}, TransTab~\cite{wang2022transtab}, PORTAL~\cite{spinaci2024portal}, and CARTE~\cite{kimcarte}.
Xtab is pretrained using what they refer to as \textit{featurizers} to generate column embeddings but it ignores the column names. CARTE leverages graphs to model multiple tables as input and is pretrained with knowledge graphs. PORTAL, TransTab, and UniTabE incorporate specific preprocessing techniques to generate feature embeddings, enriching them with column name embeddings (just as a  positional encoding would do).
TransTab is interesting because it was designed for a weak form of \textit{cross-table shift} where train and deployment columns partially overlap. This approach addresses the limitation of most pretrained models which assume a fixed table structure between training and deployment.
Research has also explored adapting LLMs for tabular data. In this context, GreaT~\cite{great23} and TabuLa-8B were introduced with distinct objectives. Both approaches convert table rows into text and finetune a pretrained LLM—GreaT for tabular data generation and TabuLa-8B for predictive tasks.
To address the scarcity of labeled data, STab~\cite{hajiramezanali2022stab} introduces a data-augmentation-free method for learning tabular data representations. STUNT~\cite{nam2023stunt} generates few-shot tasks by randomly selecting table columns as targets, leveraging a meta-learning approach for generalization.

Other methodologies have also been explored. PTaRL~\cite{PtarlYe2024}, for instance, employs prototype-based learning, while SwitchTab~\cite{switchTabJing2024} utilizes autoencoders with mutual and salient embeddings in the feature space. Additionally, TabPFN focuses on training with synthetic data to enhance deployment efficiency on small to medium-sized tables.

While most of these approaches rely on table-agnostic preprocessing and cross-table pretraining, they typically require a fixed structure during inference and they were designed to be finetuned on a specific table structure. This can be understood given the motivations behind these foundation models: to propose an encoder trained on various table structures and tasks that can later be used for prediction on a specific table.

\begin{figure}
    \centering
    \includegraphics[width=.6\textwidth]{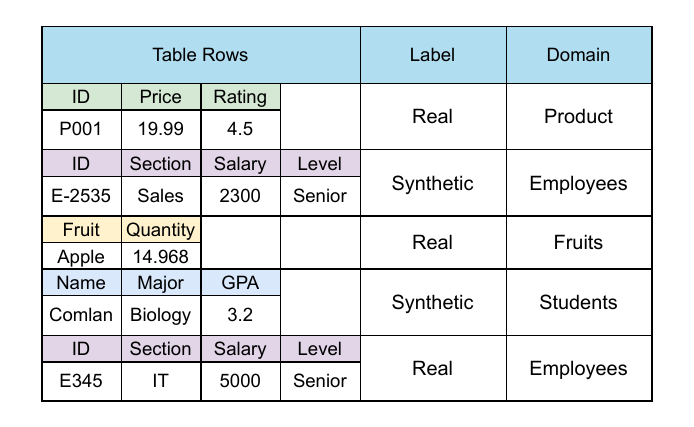}
    \caption{Example of mixed table classification instances with their labels and domains.}
    \label{fig:cross_table_example}
\end{figure}

This technical aspect represents a major challenge in adapting these approaches to the specificity of our task. We are not only requiring a \textit{table-agnostic} encoder that can be finetuned and deployed separately on different tables, we are indeed conducting what can be termed as \textit{mixed-tables} classification where the model is both trained and deployed on a mixture of different tables as it is illustrated in Figures~\ref{fig:cross_table_example} and~\ref{fig:crosstable}. Adapting the finetuning and deployment phases of these methods to our mixed-tables setup necessitates some non-trivial code rewriting that implies some undocumented choices and deviations from their initial purposes and protocols.

To address our tabular detection problem "in the wild," we introduce our \textit{Datum-wise Transformer}, which takes table rows as input during training, can be deployed for a large variety of table structures, encodes all features as text with an independent embedding for each feature, and produces column permutation-invariant results. None of the presented related methods fully meet these requirements, but we believe that the closest competitor is TaBERT, as it provides a pretrained full-table embedding that relies on a row linearization with tuples of the form \texttt{<column>|<type>|<value>}, which is similar to ours.
With minimal adaptations, this model can be deployed according to our protocol, enabling the generation of embeddings for rows from any table, regardless of its structure. Technically, adapting the TransTab implementation also appears feasible. However, its performance would likely be suboptimal, as it relies on incremental training where test tables share some columns with the training ones (e.g., patient records with additional columns introduced after model deployment). This contrasts significantly with our setup, in which test tables are entirely distinct from training ones. The same limitation applies to their zero-shot approach, where the model performs better when there is an overlap of shared columns between the training and test tables.

Given the challenges of adapting existing models to our protocol, we included BART~\cite{lewis-etal-2020-bart} as an additional pretrained baseline for evaluation. Thanks to its bidirectional encoder and pretraining procedure (involving masked tokens and denoising), BART effectively captures semantic information from input texts. This characteristic suggests it could be a relevant choice for processing our textual table rows. Similar to TaBERT, BART can be utilized within our protocol to generate embeddings for any table row.
We also included the \textit{Flat Text} transformer baseline introduced in \cite{KindjiIDA2025}. This transformer model is based on a lightweight BERT-like text encoder but it is trained from scratch.

Additionally, we evaluated PORTAL under a different protocol, where the model was trained and deployed separately for each table in our test set (table-specific finetuning and deployment). While it performed well in this context, we did not include it as a baseline for \textit{cross-table shift}, as adapting it to our protocol would require extensive code refactoring and tuning.

We provide more details about the \textit{Flat Text} approach of  \cite{KindjiIDA2025} in the next section as it  serves as a foundation for our proposed \textit{Datum-wise Transformer}.

\section{Table-agnostic Datum-wise Transformer}
\label{s:table_agnostic_eq_transf}

As previously mentioned, distinguishing between fake or generated data and real data can be framed as a binary classification task. However, in tabular data, especially for "in the wild" detection (i.e., on unseen tables), the primary challenge does not stem from the classifier itself—since any standard classifier can be used—but rather from how the data are represented before being fed into the classifier (or a classifier head in a deep learning setup). In the following, we describe our method for detecting synthetic tabular data. We begin with our \textit{datum-wise} architecture (Section~\ref{s:our_datum_wise_model}), which is designed to generate effective representations for our fake data detection task. This is followed by a description of its extension with a domain adaptation strategy aimed at further enhancing its effectiveness (Section~\ref{sec:domadapt}).

\subsection{Datum-Wise Transformer Architecture}
\label{s:our_datum_wise_model}

\begin{figure}
    \centering
    \includegraphics[width=\textwidth]{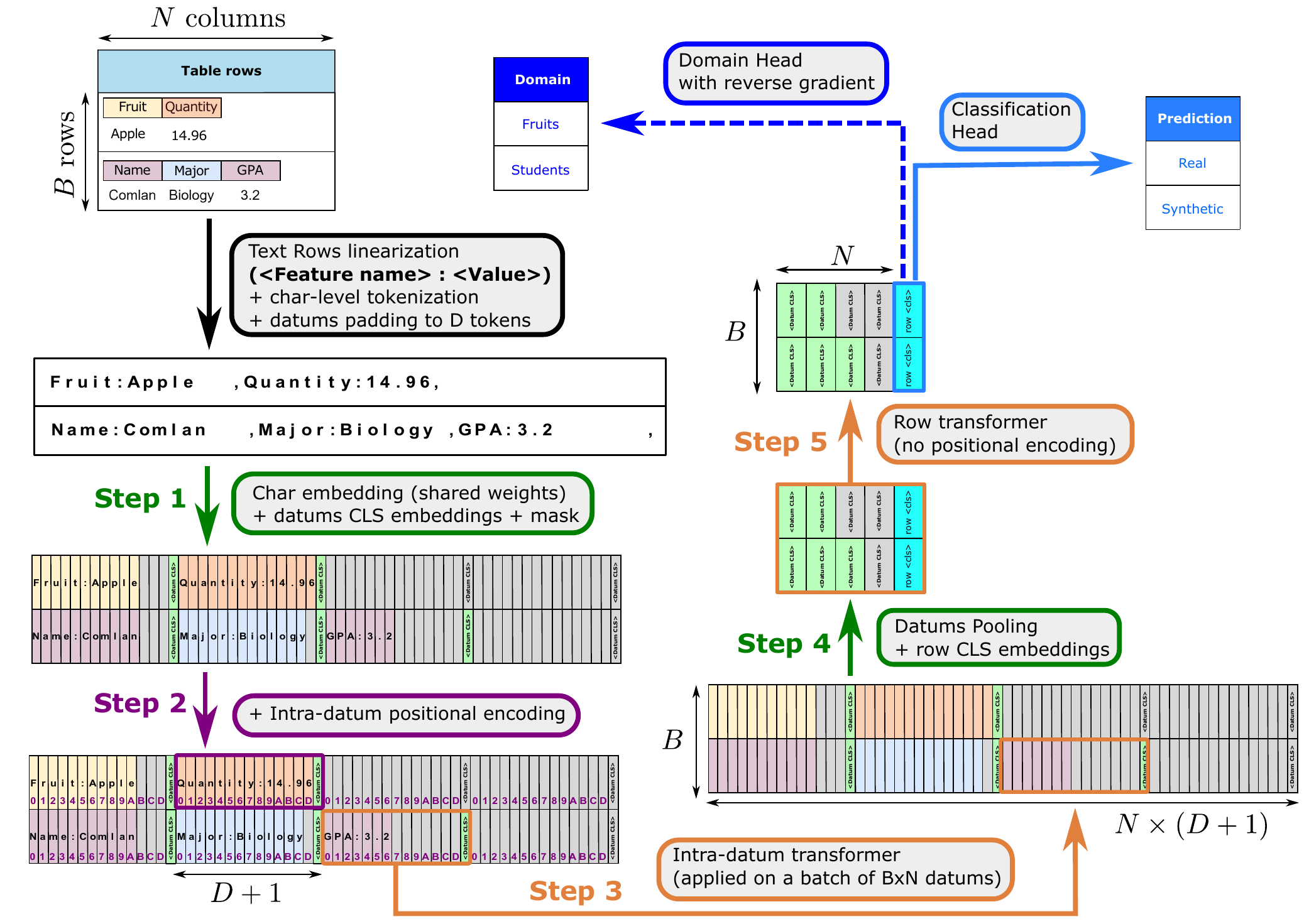}
    \caption{Datum-wise transformer pipeline with domain adaptation head.}
    \label{fig:equivarch}
\end{figure}

Our detector uses two transformers as its backbone: a \textit{datum transformer} and a \textit{row transformer}. The \textit{datum transformer} process batches of text \textit{datums} and the \textit{row transformer} works on a pooled \textit{datum} representation. The whole pipeline and architecture is described in Figure~\ref{fig:equivarch}.

Each table row is converted into a string which is the concatenation of \texttt{<column>:<value>} strings (that we call \texttt{datums}), and tokenized at a character level. %
In~\cite{KindjiIDA2025}, once the data is encoded as text, a text transformer classifier with a character-level embedding and a global positional encoding (on the entire row, as done in~\cite{great23}) is used to solve the fake/real data binary classification problem. %
The key improvement of our approach against traditional BERT-like or BERT-based text encoders such as \textit{Flat Text} or TaBERT is to restrict the positional encoding to the \textit{datums} (Step 2 in Figure~\ref{fig:equivarch}) and process them independently as a first stage.
This independent "featurization" stage is similar to the one proposed in PORTAL, TransTab, or UniTabE. However, unlike these models, which use a partially handcrafted approach based on data types, our feature encoding is inferred directly from raw data. %
At the end of this stage each \textit{datum} is encoded as a single $192$-dimensional embedding which aggregates both column name and value, thus allowing the second stage of our detector to be invariant by column permutation. By processing data at the \textit{datum} level rather than entire rows (as done for existing baselines~\cite{KindjiIDA2025}), our approach shortens the sequence length fed to the transformer detector, reducing computational costs. %
We also avoid the reliance on column order that can be induced by a global positional encoding. This can lead to significant problems if the detector is deployed on tables with a different column arrangement. The detector could easily be misled by simply inverting the order of the columns, even when presented with a table that it encountered during training.

The positional encoding within the \textit{datum transformer} (Step 2) enables it to focus on column-related information without being conditioned on any specific column order. Without it, the transformer would not distinguish the positions of the characters and the entire row would be viewed as a bag of characters (where for instance $elbow:201.1$ and $below:1.012$ would be considered identical). On the other hand, and even though the positional encoding is important within a \textit{datum}, it is a desirable property, when dealing with tabular data, and especially to detect fake data, to provide predictions that are invariant to column permutations of the table.

Technically, our model applies two levels of padding. \textit{Intra-datum} padding extends the length of each \textit{datum} block to match the longest \texttt{<column>:<value>} string. Then, \textit{extra-datum} padding adds dummy datum blocks to handle varying numbers of \textit{datums} in each table of the training set. Each \textit{datum} is appended with a \textit{CLS} token, serving as a representation of the feature. In the following sections, we refer to these tokens as \textit{CLS-Datums}. 

The results from the character embedding step and the application of positional encoding (in Step 2 of Figure~\ref{fig:equivarch}) are input into the \textit{datum transformer} (Step 3) and we only retrieve the \textit{CLS-Datums} embeddings from its output (Step 4). This datum pooling operation reduces drastically  the input size for the subsequent steps. We expect the \textit{CLS-Datums} to provide sufficient information to effectively represent the attributes. Intermediate levels of pooling could also be used to retrieve additional tokens as \textit{datum} representations alongside the \textit{CLS-Datums}. While we did not evaluate this approach in our current work, we leave it for future exploration. 

The \textit{CLS-Datums} are appended with an additional row-level \textit{CLS} token (192-dimensional, like the \textit{CLS-Datums}) that will serve for our classification task. In the following paragraphs, we will refer to this token as \textit{CLS-Target}. The result of this operation is input to the \textit{row transformer} (Step 5). This transformer does not incorporate any positional encoding as all the position-related information are already processed by the \textit{datum transformer}. We extract the \textit{CLS-Target} from the output and pass it to the classification head. Our detector is trained using a binary cross entropy loss.

\subsection{Domain Adaptation}
\label{sec:domadapt}

To enhance our detector's performance "in the wild", we integrate a well-established domain adaptation strategy that improves a detector's ability to generalize from a specific data distribution to unlabeled target data with a different distribution. Specifically, we employ the gradient reversal techniques from~\cite{ganin2015unsupervised,Saito_2018_ECCV} to minimize the classifier's reliance on table structures (with table names considered as "domains") in its embeddings while emphasizing the values within the cells of the tables.
For example, some tables may exhibit characteristics that make them easily distinguishable from others, such as unique data distributions: a table containing only numerical values versus one with a mix of text and numbers. To operate "in the wild", the detector must learn to differentiate synthetic rows from real ones without relying on table-related features. 

In practice, we add a domain classification head in our architecture from which the gradient reversal will be applied down to the representation learning layers. This domain classification head also utilizes the \textit{CLS-Target} produced by the \textit{row transformer} for its predictions. This is shown in the upper right part of the Figure~\ref{fig:equivarch}. We still use a cross-entropy loss for optimization.

\section{Experiments}
\label{s:experiments}

We first outline our experimental protocol for evaluating our \textit{datum-wise} transformer for synthetic tabular data detection "in the wild" before diving into the experimental results which compare our detector to the \textit{Flat Text} model, and to the embeddings from BART and TaBERT (referred to as "BART-embd" and "TaBERT-embd") discussed in Section \ref{s:related_work}.

\subsection{Training Data}
\label{s:training_data}

\begin{table}[t!]
\small
\centering
    \caption{Description of the tables considered in the experiments. "Size" is the number of total instances in the table, "\#Num" refers to the number of numerical attributes and "\#Cat" the number of categorical ones for each table.}
    \label{tab:datasets}
    \centering
    \begin{tabular}{cccc}\hline
        Name & Size & \#Num &  \#Cat\\\hline
        Abalone\tablefootnote{\label{fn:dataset_link_openml}\url{https://www.openml.org}} & 4177 & 7 & 2\\
        Adult\footref{fn:dataset_link_openml} & 48842 & 6 & 9 \\
        Bank Marketing\footref{fn:dataset_link_openml} & 45211 & 7 & 10 \\
        Black Friday\footref{fn:dataset_link_openml} & 166821 & 6 & 4 \\
        Bike Sharing\footref{fn:dataset_link_openml} & 17379 & 9 & 4 \\
        Cardio\tablefootnote{\label{fn:dataset_link_kaggle}\url{https://www.kaggle.com/datasets}} & 70000 & 11 & 1 \\
        Churn Modelling\footref{fn:dataset_link_kaggle} & 4999 & 8 & 4 \\
        Diamonds\footref{fn:dataset_link_openml} & 26970 & 7 & 3 \\
        HELOC\footref{fn:dataset_link_kaggle} & 5229 & 23 & 1 \\
        Higgs\footref{fn:dataset_link_openml} & 98050 & 28 & 1 \\
        House 16H\footref{fn:dataset_link_openml} & 22784 & 17 & 0 \\
        Insurance\footref{fn:dataset_link_kaggle} & 1338 & 4 & 3 \\
        King\footref{fn:dataset_link_kaggle} & 21613 & 19 & 1 \\
        MiniBooNE\footref{fn:dataset_link_openml} & 130064 & 50 & 1 \\\hline
    \end{tabular}
\end{table}

We use the same tables as~\cite{KindjiIDA2025} for comparability (see Table~\ref{tab:datasets}). The training data consists of a mix of real and synthetic rows for each table, where a row is considered synthetic if generated by one of the data generators. The state-of-the-art tabular generator models considered are: TabDDPM~\cite{kotelnikov2023tabddpm}, TabSyn~\cite{zhang2023mixed}, TVAE, and CTGAN~\cite{CTGAN}. They were all heavily tuned following the protocol proposed in \cite{kindji2024hoodtabulardatageneration}. To ensure consistency, we use the exact synthetic tables provided by~\cite{KindjiIDA2025} without generating new ones. The detectors are evaluated on a 3 folds cross-validation setup.

The final training and testing datasets are constructed as shown in Figure~\ref{fig:crosstable}. The synthetic rows for each table consist of an equal number of rows sampled from each generator. Moreover, our setup follows a realistic protocol in which the generator has access to all available real data. This contrasts with the C2ST method, where the generator is limited to a subset of the training data.

\subsection{Baselines}
\label{s:baselines}

\subsubsection{Flat Text Transformer~\cite{KindjiIDA2025}} (See Section~\ref{s:table_agnostic_eq_transf}). The encoder of this model is trained from scratch. 
We evaluate our \textit{datum-wise} transformer against the existing \textit{Flat Text} transformer baseline.

\subsubsection{TaBERT  embedding~\cite{yin20acl}}

As stated earlier, this pretrained model is designed to encode entire tables alongside with NL sentences. Each row in our pool of table is considered as a table and encoded by a pretrained version of TaBERT (see details in Section \ref{s:related_work}). We consider the pretrained version with \textit{K=1}~\footnote{\url{https://github.com/facebookresearch/TaBERT}}, as we anticipate it to be more suitable for our setup, where we process tables on a row-by-row basis. We deploy the \textit{TaBERTbase} version initialized from \textit{BERTbase}, which has 12 heads and 12 layers of attention.

The context for each row was generated by prompting \textit{GPT-4O-mini} to describe each table in our pool (Table~\ref{tab:datasets}). The pretrained model then processes a row formatted as a table, along with an associated context, as recommended by the authors. We retrieve each row's \textit{CLS} embedding and train a classification head. We do not update TaBERT's weights due to the high cost involved; instead, we only train the classification head. As the authors noted, their model can be viewed as an encoder for systems that require table embeddings as input.

\subsubsection{BART  embedding~\cite{lewis-etal-2020-bart}} 

We use a pretrained version of BART (the \textit{bart-base} checkpoint which has 12 heads and 6 layers of attention) and train it using the same procedure as for TaBERT. We generate embeddings for each row and extract the \textit{CLS} token as input for the classification head. During training, only the weights of the classification head are updated due to the high cost involved here as well.

\subsubsection{Implementation details}

The classification heads used for BART and TaBERT includes a batch normalization layer, followed by a single linear layer and a Sigmoid activation function. Both models produce embeddings of dimension 768, which are subsequently input into their respective classification heads. These classification heads are optimized using AdamW~\cite{loshchilov2018decoupled} with a learning rate of 5e-5. The models are trained for 10 epochs, with early stopping criteria applied if there is no improvement on the validation AUC over three successive epochs. This number of epochs has been sufficient, as all models stopped training before reaching the maximum limit, with each epoch taking approximately one and a half hours.

In our \textit{datum-wise} method, both the \textit{datum transformer} and the \textit{row transformer} consist of $3$ attention layers with $6$ heads in each layer. We employ a fixed learning rate of $1e-5$ with the Adam~\cite{adamKingma2014} optimizer. The classification head follows the architecture we use for BART and TaBERT. The domain head also adopts the same architecture, using a Softmax activation function. As stated earlier, all detectors are trained with a cross entropy loss. We plan to release our source code once administrative and licensing issues are resolved and after the peer review period.

\subsection{Detection Setup}
\label{s:detection_setup}

\begin{figure}[t!]
    \centering
    \includegraphics[width=\textwidth]{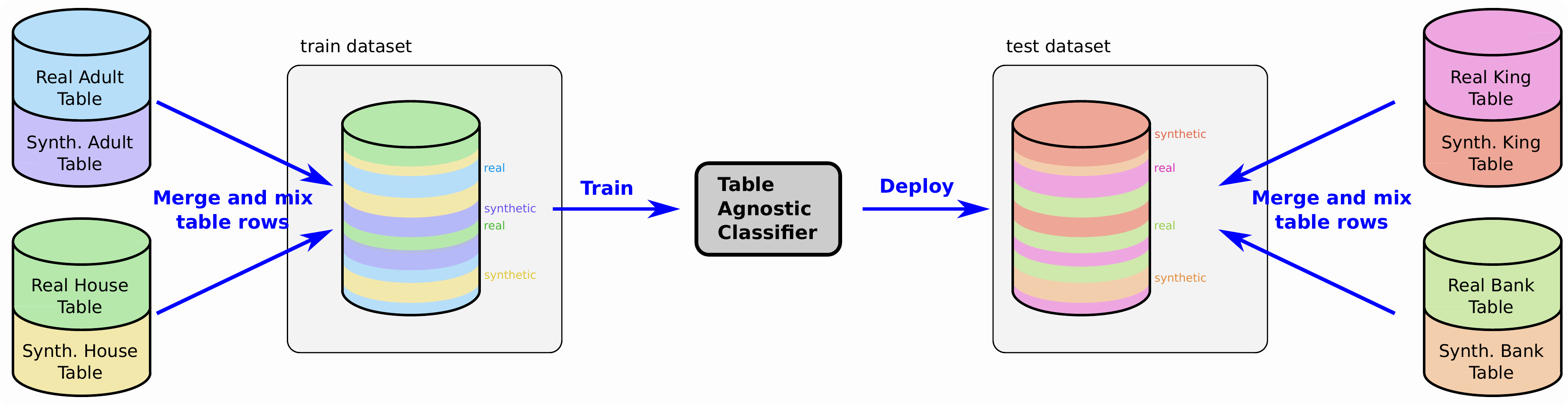}
    \caption{\textit{Cross-table shift} protocol: the real-vs-synthetic detector is trained on a mixture of table rows and tested/deployed on a mixture from holdout tables.}
    \label{fig:crosstable}
\end{figure}

All detectors are evaluated in the \textit{cross-table shift} setup, where the model is deployed on unseen tables (Figure~\ref{fig:crosstable}). It is important to note that our implementation (following~\cite{KindjiIDA2025}) involves a \textit{cross-table shift} between both the training and validation sets, as well as between the training and test sets. For our \textit{datum-wise} method, we evaluate two versions: one with domain adaptation and one without. Performance is reported using the ROC-AUC and accuracy metrics.

\subsection{Results}
\label{s:results}

The results from our experiments are reported in Table~\ref{tab:all_results}. We provide the reported performance for the considered baselines and our \textit{datum-wise} method evaluated with and without domain adaptation.

\begin{table}[t!]
\caption{AUC and Accuracy performance reported for transformer detectors. Our proposed \textit{Datum-wise} method is evaluated with and without domain adaptation (DA). "BART-embd" and "TaBERT-embd" refer respectively to the embeddings produced by BART and TaBERT.}
\label{tab:all_results}

\centering

\begin{tabular}{ccc}
\toprule
\multirow{2}{*}{Model} & \multicolumn{2}{c}{Metrics} \\
        \cline{2-3}
                                                    & AUC & Accuracy \\ 
            \midrule

          BART-embd & $0.50 \pm 0.00 $ & $0.50 \pm 0.00$ \\
        
        \cline{1-3}
          TaBERT-embd & $0.51 \pm 0.00$ & $0.50 \pm 0.00$ \\
        
        \cline{1-3}
        
         Flat Text  & $0.60 \pm 0.07$ & $0.52 \pm 0.01$ \\
         
        \cline{1-3}
         Datum-wise  & $0.67 \pm 0.05$ & $0.59 \pm 0.08$ \\

        \cline{1-3}
         Datum-wise + DA  & \textcolor{Black}{\bm{$0.69 \pm 0.04$}} & \textcolor{Black}{\bm{$0.66 \pm 0.05$}} \\
\bottomrule
\end{tabular}
\end{table}

\subsubsection{Without Domain Adaptation}

Our method consistently outperforms all baseline approaches including the \textit{Flat Text} method we built upon, across all evaluated metrics. We achieve an average AUC of 0.67 (resp. accuracy of 0.59), establishing state-of-the-art performance for the \textit{cross-table shift} detection setup. 

\begin{figure}[ht]
    \centering

    \setlength{\abovecaptionskip}{1pt}  

    \subfigure[\textit{Flat Text} Transformer~\cite{KindjiIDA2025}]
    {%
        \includegraphics[width=0.45\linewidth]{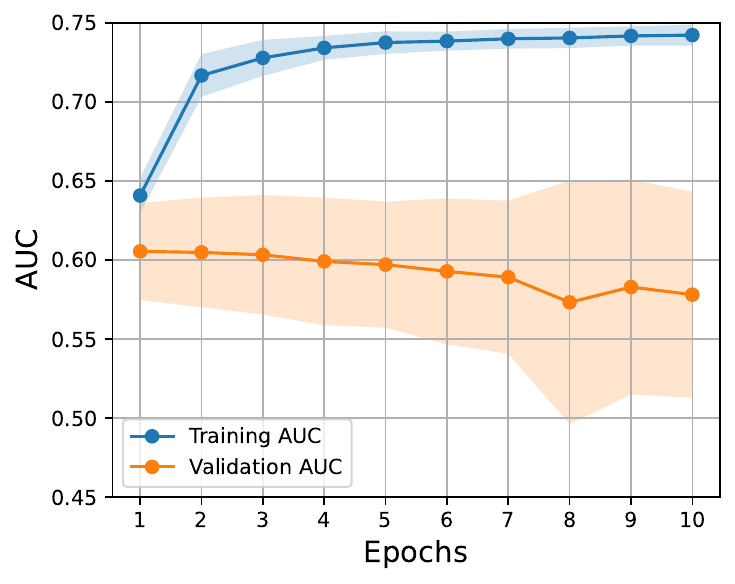}
        \label{fig:flat_text_auc_variance}
    }
    \hfill
    \subfigure[Our \textit{datum-wise} Transformer]
    {%
        \includegraphics[width=0.45\linewidth]{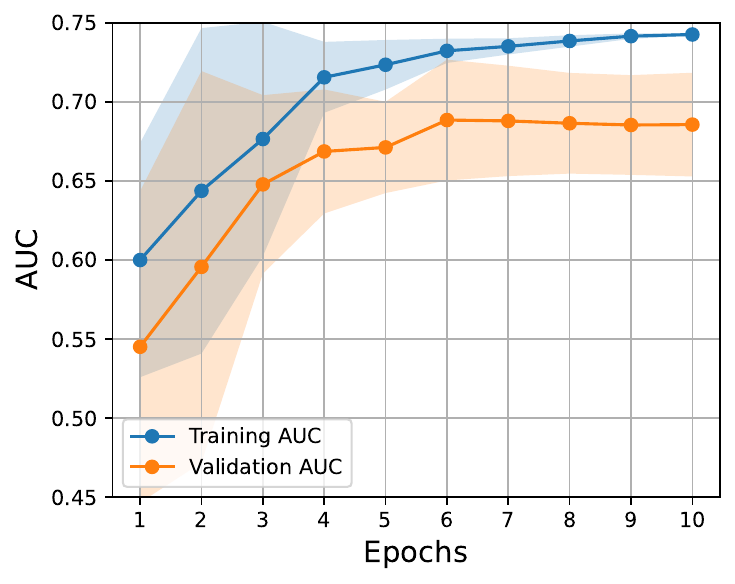}
        \label{fig:datum_auc_variance}
    }
    \caption{We present the average and standard deviation of AUC performance during the first 10 epochs of training and validation under \textit{cross-table shift}.} 
    \label{fig:auc_variance_curves}
\end{figure}

A closer examination of the training logs shows that we surpass the \textit{Flat Text} transformer baseline not only on the test set (i.e. "in the wild") but also consistently during training on the validation set. Our results show an average best AUC of $0.71$ on the validation set, compared to $0.62$ for the \textit{Flat Text} baseline, demonstrating the superiority and effectiveness of our approach. %
Note that these results are also more steady across folds, with a standard deviation of $0.01$ compared to $0.04$ for the \textit{Flat Text} baseline. This is illustrated in Figure~\ref{fig:auc_variance_curves} for both methods. The figure shows the mean and standard deviation of the AUC across the three folds for the first 10 epochs of training and validation. 

By examining Figure~\ref{fig:auc_variance_curves}, we observe that the \textit{Flat Text} transformer baseline consistently hovers around an average AUC of $0.60$ in validation, while our \textit{datum-wise} approach achieves this average performance only after two epochs. We can also notice an overlap between the standard deviation performance of our \textit{datum-wise} model during validation and training while the \textit{Flat Text} baseline consistently remains below the training performance. These results indicate that our \textit{datum-wise} method outperforms the \textit{Flat Text} method at capturing information necessary for the classification task.

As for BART and TaBERT's embeddings (respectively \textit{BART-embd} and \textit{TaBERT-embd}), we observe that they achieve performance close to random, with an AUC of $0.50$ for BART and $0.51$ for TaBERT. Both models obtain an accuracy of $0.50$. 

Analyzing the training logs reveals a notable improvement in performance on the training set, with an average AUC of $0.63$ for TaBERT's embedding and $0.59$ for BART. TaBERT's embeddings seem to provide a richer representations than BART for this task, as suggested by the training performance. However, both models struggle to generalize to the validation and test sets. Notably, the performance on the validation set remains just above $0.51$ for both models, suggesting that despite the training improvements, there are underlying challenges that hinder effective generalization.
The poor performance observed can be attributed to the fact that our usage of these models differs from their original design intent. Even with careful consideration of the specific pretrained model for TaBERT (\textit{K=1}), the evaluation results remain unremarkable.

\subsubsection{With Domain Adaptation}
In our domain adaptation setup, the \textit{lambda} parameter controls the strength of the gradients propagated back from the domain classification head. This parameter is gradually increased from 0 to 1 over the course of training. Initially, the model requires a few iterations to learn the primary target classification task (distinguishing between real and synthetic data) before being exposed to negative gradients. This delay prevents it from prematurely focusing on table-related features, which are learned through the domain classification head (i.e., distinguishing between table names).

An important insight emerged when we first experimented with the original scheduling approach proposed by~\cite{ganin2015unsupervised}. We found that this schedule was too aggressive for our setup, often causing early stopping after just a few epochs. To address this issue, we tested a \textit{smoother} cosine-based \textit{lambda} schedule, which ultimately delivered the best performance, as shown in Table~\ref{tab:all_results}. With this domain adaptation strategy, accuracy improved from $0.59$ to $0.66$, and AUC increased from $0.67$ to $0.69$. The simultaneous improvement in both metrics suggests that the model was initially influenced by table-related features, but this effect was mitigated through the adapted training strategy.

\section{Conclusion, Limitations, and Future Work}
\label{s:conclusion}
In this work, we introduce a \textit{datum-wise} transformer architecture designed for detecting synthetic tabular data “in the wild,” specifically in a \textit{cross-table shift} setting, where the model encounters table structures it has never seen before. We propose a transformer trained on character-level embeddings with a local positional encoding applied at the column level. We compare to existing baselines~\cite{KindjiIDA2025} and we show that our architecture consistently outperforms all baselines across all metrics. Furthermore, we enhanced the model by integrating a domain adaptation strategy.

{
Regarding the experiment results, we were surprised by the poor performance of the pretrained encoders TaBERT and BART. We plan, on short schedule, to finetune these encoders more thoroughly on our task and test domain adaptation to see if it improves.}

Our architecture opens several promising avenues for future research. First, it can be leveraged in a pretraining-finetuning framework for predictive tasks on tabular data. This could involve various pretraining objectives inspired by existing work, such as Masked Language Modeling (MLM) in TaBERT or few-shot learning techniques like STUNT. Second, we can build upon prior work such as TransTab to address the fixed-table structure assumption inherent in most pretrained models. Our work can be extended to \textit{mixed-tables} classification and regression downstream tasks, involving training and deploying on a mixture of different tables.

\begin{credits}


\end{credits}
%
%
%
\bibliographystyle{splncs04}
\bibliography{biblioCharbel}

\end{document}